\documentclass[]{article}

\usepackage[margin=0.7in]{geometry}
\usepackage{authblk}
\usepackage{algorithmic}
\usepackage{algorithm}
\usepackage{color}
\usepackage{multirow}
\usepackage{amsmath}
\usepackage{subfigure}
\usepackage{verbatim}
\usepackage{array}
\usepackage[super]{natbib}
\bibpunct{}{}{,}{s}{,}{,}

  \usepackage{amsmath, amssymb, amstext, amsthm}

	\usepackage{graphicx}

\title{A Weakly Supervised Learning Approach\\ based on Spectral Graph-Theoretic Grouping}
\date{}
\author[1,2]{Tameem Adel*}
\author[2]{Alexander Wong}
\author[2]{Daniel Stashuk}
\affil[1]{Institute for Computing and Information Science (iCIS) of the Faculty of Sciences, Radboud University, Nijmegen, The Netherlands}
\affil[2]{Systems Design Engineering, University of Waterloo, Waterloo, ON, Canada}
\begin{document}

\maketitle

\small

\begin{abstract}
In this study, a spectral graph-theoretic grouping strategy for weakly supervised classification is introduced, where a limited number of labelled samples and a larger set of unlabelled samples are used to construct a larger annotated training set composed of strongly labelled and weakly labelled samples. The inherent relationship between the set of strongly labelled samples and the set of unlabelled samples is established via spectral grouping, with the unlabelled samples subsequently weakly annotated based on the strongly labelled samples within the associated spectral groups.  A number of similarity graph models for spectral grouping, including two new similarity graph models introduced in this study, are explored to investigate their performance in the context of weakly supervised classification in handling different types of data.  Experimental results using benchmark datasets as well as real EMG datasets demonstrate that the proposed approach to weakly supervised classification can provide noticeable improvements in classification performance, and that the proposed similarity graph models can lead to ultimate learning results that are either better than or on a par with existing similarity graph models in the context of spectral grouping for weakly supervised classification.
\end{abstract}

\begin{comment}
\begin{abstract}
A spectral graph-theoretic grouping strategy for weakly supervised classification is introduced where a limited number of labelled samples are used to construct a larger training set (composed of a mix of labelled samples and associated unlabelled samples). The objective is to improve classification results. Weakly annotated data are labelled by constructing similarity graphs that are more flexible in handling different densities in different datasets as well as different densities within a dataset due to the introduction of adaptable parameters. Experimental results demonstrate that the proposed similarity graphs lead to ultimate learning results that are either better than or on a par with the other similarity graph construction techniques.
\end{abstract}
\end{comment}

%\input{introduction}

%\input{related_work}

%\input{results}

%\input{discussion}

%\input{Methods}

%%%%%%%%%%%%%%%%%%%%%%%%%%%%%%%%%%%%%%%%%%%%%%%%%%%%%%%%%%%%%%%%%%%%%%%%%%%%%%%%%%%%%%%%%%%%%%%%%%%%%%%%%%%%%%%%%%%%%%%%%%
%\section{Introduction}\label{s:introduction}

\section{Introduction}\label{s:introduction}
In weakly supervised learning paradigms, a learner receives a training set consisting of a limited amount of labelled data as well as a fairly larger amount of unlabelled data. Without loss of generality, assume that we have a dataset consisting of bags of instances with binary labels that are assigned based on the existence of a specific object of interest. Each bag consists of instances. Bags, as well as instances can be labelled positive or negative. A positive label means that the corresponding bag (instance) contains (is) the object, whereas a negative label signifies that the bag (instance) does not contain (is not) the object. Labels are provided for the bags only. If a bag is labelled negative, then it is fully labelled as all the instances of the bag definitely do not represent the object of interest, and therefore all the instances or parts of the bag are labelled. A positive-labelled bag is considered fully labelled if the instances representing the object of interest are indicated in the bag. A bag can be positive-labelled but no information is provided about which instances represent the object in the bag or, put in other words, instance labels are treated as latent variables. Datasets containing the latter kind of bag represent an example of datasets handled by weakly supervised learning paradigms. With all instances of positive-labelled bags being unlabelled, performance of a classifier will undoubtedly be far from optimal. The aim of the weakly supervised learning approach introduced in this study, is to weakly annotate the unlabelled instances so that performance of a classifier applied subsequently on the data can be improved. Weakly supervised learning paradigms are recently gaining further significance in machine learning due to the increasing importance of learning from unlabelled data. Obtaining fully labelled data is apparently very useful but is error-prone and, more importantly, expensive to obtain \citep{Galleguillos08}, and this is why weakly supervised learning paradigms are crucial.

%you could also talk about pixels rather than bounding boxes as you can see in "weakly labelled training examples indicate which objects of interest are present in training images without specifying the pixels that are associated with them." in %galleguillos_eccv08_0.pdf

A real-world challenge that lends very well to the application of weakly supervised learning paradigms where datasets have a bags-of-instances form is the problem of muscle classification based on electromyographic (EMG) signals.  Here, the datasets consist of sets of electromyographic (EMG) signals that represent a particular muscle, with each EMG signal comprising of EMG signal contributions from the different components that make up the muscle. Each component of a muscle is referred to as a motor unit (MU) and the EMG contribution of each component is referred to as a motor unit potential train (MUPT). Each MU has one label out of three; normal, myopathic or neurogenic. The same three labels are used to label each muscle. In the training dataset, there are feature values for each component of the muscle, i.e.\ for each MU, while labels are available only for each muscle as a whole. For two of the three labels (myopathic and neurogenic, and it will be referred to both of them together here as disordered), a muscle that is ``myopathic'' or ``neurogenic'' contains both normal and myopathic or neurogenic components, respectively. On the other hand, in a ``normal'' muscle, all components are normal. This can be seen as a multiclass version of the bags-of-instances example described earlier. Rather than having two labels; one containing fully labelled instances and another containing unlabelled instances, there are three labels, one containing fully labelled instances and two containing unlabelled instances. Instances of normal muscles are fully labelled because they are all labelled normal, while instances of both myopathic and neurogenic muscles are unlabelled because it is not known which of them are disordered and which are normal. Therefore, the normal label in muscles is equivalent to the negative label in bags while both the myopathic and neurogenic labels in muscles are equivalent to the positive label in bags.

There are numerous examples of learning algorithms where unlabelled or weakly labelled data are utilised \citep{Arora07, Chum07, Lee11, Winn05, Crandall06}. Also, \citet{Bergamo10} provide another example where they exploit weakly annotated web images to build a weakly supervised object classifier. In addition to object recognition in images, weakly supervised learning has other applications in computer vision. For example, \citet{Prest12} perform learning based on weakly labelled videos and fully labelled images in order to detect objects from web videos. Other examples of weakly supervised learning algorithms that are applied on videos include \citet{Ali11} and \citet{Leistner11}.

Multiple-instance learning (MIL) algorithms are closely related to weakly supervised data. In MIL, training instances are grouped together in bags. Each bag has a label associated to it. Each instance belongs to only one bag. Each instance has a label associated to it that can be the same or different from the label of the bag it belongs to. Instance labels are not observed.
%\citet{Tam13} exploit Bayes nets to construct MIL generative models that target a solution for the muscle classification problem.
\citet{Galleguillos08} train an MIL discriminative classifier on weakly annotated image data. \citet{Andrews03} represents another example of an MIL algorithm applied on weakly annotated data, while \citet{Blaschko10}, \citet{Vezhnevets10} and \citet{Vijayanarasimhan08} are examples of MIL-based approaches applied on weakly labelled images for the purpose of object recognition.

The main objective of this study is to turn a training dataset consisting of a limited number of labelled instances and a larger number of unlabelled instances into a larger annotated training dataset consisting of weakly labelled instances, which are those that were unlabelled, and strongly labelled instances, for the sake of improving classification performance. This is to be achieved via the proposed weakly supervised learning approach. Unlabelled data are weakly annotated by applying a spectral graph-theoretic grouping strategy that makes use of strongly labelled instances as well as similarity among instances in order to assign weak labels to the unlabelled instances. Spectral graph-theoretic grouping is based on similarity graph models. In addition to the similarity graph models in the literature, two new similarity graph models are introduced in this study. Weakly labelled and strongly labelled instances form a larger annotated training dataset. By utilizing the proposed spectral grouping strategy to facilitate for weakly supervised learning, one can obtain a larger annotated training dataset composed of strongly and weakly labelled data that should consequently lead to an improved classification performance compared to the case when only strongly labelled data, or strongly labelled data and unlabelled data, are used. This study serves as a methodological guide to other weakly supervised learning paradigms, and is not limited to the aforementioned EMG muscle classification example. In addition to being a part of the proposed weakly supervised learning paradigm, performance of the new similarity graph models is tested separately on three benchmark datasets, Abalone, Swiss Banknotes and Segmentation. The Abalone dataset consists of abalone's physical measurements (input), and it is required to predict the abalone's age (output) from these measurements. Features of the Swiss Banknotes dataset are explanatory variables describing characteristics of Swiss banknotes and the goal is to decide whether or not each banknote is genuine \citep{Flury83}. The Segmentation dataset contains images, each representing one out of seven outdoor objects, i.e.\ grass, path, window, cement, foliage, sky or brickface \citep{Ung05}.

\begin{figure}[!]
 \centerline{\includegraphics[width=175mm,height=150mm]{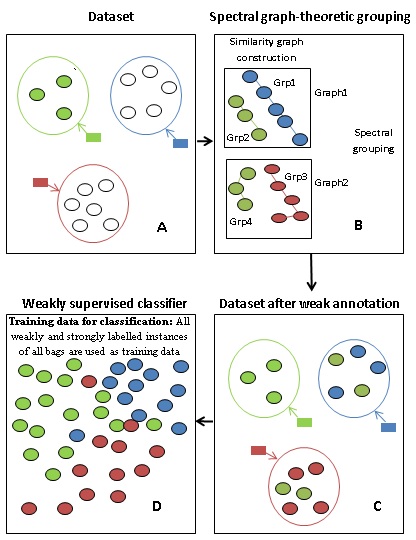}}
  \caption[A schematic representation of the main steps of the proposed weakly supervised learning approach]{A schematic representation of the main steps of the proposed weakly supervised learning approach. \textbf{A.} A dataset of the bags-of-instances setting (only one bag of each label is shown for simplicity but cardinality of bags of each label is greater). Each bag label is represented by a colour. \textbf{B.} All instances of blue bags are grouped in one similarity graph model (Graph1) and the same for instances of red bags (Graph2). Spectral graph-theoretic grouping is performed on each similarity graph model to group the instances in two groups. Relation with the green labelled instances decide the label of each group. In this example, the assumption is that across all blue (resp.\ red) bags, the total number of green instances is less than the total number of blue (resp.\ red) instances. Thus, the group with greater cardinality is assigned the blue (resp.\ red) label. Therefore, Grp1 is assigned the blue label, Grp3 is assigned the red label and both Grp2 and Grp4 are assigned the green label. \textbf{C.} Now instances of blue and red bags are weakly annotated while instances of green bags are strongly labelled. \textbf{D.} All weakly annotated instances, i.e.\ instances of all blue and red bags, as well as all strongly labelled instances, i.e.\ instances of all green bags, are used as one larger set of training data by a weakly supervised classifier.\label{fig:one}}
\end{figure}

The main contribution of this study is the introduction of a new weakly supervised learning approach based on spectral graph-theoretic grouping. The approach mainly targets datasets of the bags-of-instances form, like the one shown in Figure~\ref{fig:one}(A). Instances belonging to the green labelled bags (e.g.\ the bag at top left) are labelled, but instances belonging to blue and red (e.g.\ the other two) bags are unlabelled. For instances of each of the latter two bag labels, instances are grouped together in one spectral graph. More specifically, all instances of blue labelled bags are grouped together in one similarity graph, which is Graph1 in Figure~\ref{fig:one}(B) whereas Graph2 in Figure~\ref{fig:one}(B) contains all instances of red labelled bags. Spectral graph-theoretic grouping is performed by first constructing similarity graph models and then performing spectral grouping (Figure~\ref{fig:one}(B)). As per this step, two similarity graph models are proposed. Using the groups resulting from spectral grouping along with the strongly labelled instances associated within the spectral groups, unlabelled instances are weakly annotated and the result is Figure~\ref{fig:one}(C). The assumption is that across all blue (resp.\ red) bags, the total number of green instances is less than the total number of blue (resp.\ red) instances. Thus, the group with greater cardinality is assigned the blue (resp.\ red) label. The premise is that by applying an efficient grouping strategy on nodes of each similarity graph, we can weakly (but reliably) annotate the involved instances. By doing so, we construct a larger annotated training set. Finally a weakly supervised classifier exploits the whole dataset consisting of strongly labelled data and weakly labelled data (Figure~\ref{fig:one}(D)). In this study, we will not only introduce a new weakly supervised learning approach, but will also investigate and devise a number of similarity graph models and study their effect on the ability to obtain reliable weak annotations for unlabelled instances.

%%%%%%%%%%%%%%%%%%%%%%%%%%%%%%%%%%%%%%%%%%%%%%%%%%%%%%%%%%%%%%%%%%%%%%%%%%%%%%%%%%%%%%%%%%%%%%%%%%%%%%%%%%%%%%%%%%%%%%%%%%
%\section{Related work}\label{s:related_work}

%%%%%%%%%%%%%%%%%%%%%%%%%%%%%%%%%%%%%%%%%%%%%%%%%%%%%%%%%%%%%%%%%%%%%%%%%%%%%%%%%%%%%%%%%%%%%%%%%%%%%%%%%%%%%%%%%%%%%%%%%%
\section{Methods}\label{sec:Methods}
Graph-theoretic grouping has been studied before in the literature for a number of different applications. \citet{Fowlkes04} used spectral graph-theoretic grouping in an image segmentation application. They developed spectral groups which were based on using a small number of samples and extrapolating so that the computational requirements are reduced. \citet{Aksoy99} developed a graph-theoretic grouping algorithm that was used for image grouping. They grouped images based on the observation that visually similar images are also similar in the feature space because they have similar feature vectors. \citet{Wu93} represent one further example of a graph-theoretic grouping algorithm where they develop an algorithm for image segmentation. They performed grouping by building an undirected graph using data instances, then forming mutually exclusive subgraphs by gradually removing arcs according to a certain criterion.  Here, we make use of graph-theoretic grouping for a completely different purpose: for improved weakly annotation of unlabelled data for weakly supervised learning.

One strategy of weakly annotating unlabelled data is to apply a grouping technique so that each part of the unlabelled data can be related to a group, which in turn is assigned a certain label depending on the inherent relationships between the strongly labelled part of the data and the unlabelled part. Due to the fact that spectral grouping acts only on the unlabelled part of the data, the introduced spectral graph-theoretic grouping strategies are applicable on fully unlabelled datasets because they practically act on the unlabelled part of the data, as long as other grouping issues (e.g. number of groups) can be handled via the problem assumptions. For the EMG datasets, instances belonging to each disordered type of bag (myopathic or neurogenic) are known to be either normal or of the same disorder. Therefore, there are two groups. For the benchmark datasets used, the number of groups is known. As a preliminary phase of this study, spectral grouping as well as other grouping algorithms were applied to the EMG data. Normalised spectral graph-theoretic grouping according to Shi and Malik \citep{Shi2000} performed slightly better than other spectral graph-theoretic grouping algorithms which in turn performed better than other grouping algorithms. However, the improvement provided by Shi and Malik's normalised spectral graph-theoretic grouping was not considerable. After a careful inspection of the reason why Shi and Malik's normalised spectral graph-theoretic grouping does not perform better than it does, it turned out to be the fact that all similarity graph models used before in the spectral clustering literature do not capture well the pairwise similarities between instances. It is worth noting that there are two main normalised spectral graph-theoretic grouping algorithms; one is according to Shi and Malik \citep{Shi2000} and the other is according to Ng et al.\ \citep{Ng2002}. For the sake of simplicity, the former is shortly referred to in this study as normalised spectral graph-theoretic grouping, unless stated otherwise.

\subsection*{Similarity Graph Models}
The first step of the proposed weakly supervised learning approach is to perform spectral graph-theoretic grouping on the unlabelled part of the dataset. Spectral graph-theoretic grouping in turn begins by forming similarity graph model(s) of unlabelled data. In the literature, there are several popular similarity graph models that transform a given set $x_1, . . . , x_n$ of data instances with pairwise similarities $s_{ij}$ or pairwise distances $d_{ij}$ into a graph. When constructing a similarity graph model the goal is to model the local neighbourhood relationships between data instances. The following is a list of the main similarity graph models in the literature.

\textbf{$\epsilon$-Neighbourhood Graph:} Instances that have pairwise distances among each other less than $\epsilon$ are connected while the instances with pairwise distances greater than or equal to $\epsilon$ are not. Weights are considered to be at the same scale of distances; which is at most $\epsilon$ \citep{Luxburg07}. Therefore an $\epsilon$-neighbourhood graph is unweighted.

\textbf{$k$-Nearest Neighbour Graph}: An instance $x_i$ is connected to an instance $x_j$ if $x_j$ is among the $k$-nearest neighbours of $x_i$. The neighbourhood relationship is not symmetric and due to that the resulting graph is a directed graph. Therefore, the graph should be transformed into an undirected graph. One way of transforming it into an undirected graph is by connecting two instances $x_i$ and $x_j$ if $x_j$ is among the neighbours of $x_i$ ``or'' $x_i$ is among the neighbours of $x_j$. The resulting undirected graph is referred to as the $k$-nearest neighbour graph (or symmetric $k$-nearest neighbour graph \citep{Maier07}). Another way is to connect two instances $x_i$ and $x_j$ if $x_j$ is among the neighbours of $x_i$ ``and'' $x_i$ is among the neighbours of $x_j$. The resulting undirected graph in this case is referred to as the mutual $k$-nearest neighbour graph. After building the similarity graph model in both cases, edges of the graph are weighted by measuring the similarity of the respective vertices \citep{Luxburg07}.

\textbf{Fully Connected Graph:} All instances are connected to one another; in other words all instances are considered ``similar'' to one another. The edges are weighted by $s_{ij}$. The graph is useful only when local neighbourhoods can be modelled by the similarity function because this is the only way by which the fully connected graph can represent the local neighbourhood relationships \citep{Luxburg07}. The Gaussian similarity function \begin{math}s(x_i, x_j) = exp(-||x_i-x_j||^2/(2\sigma^2))\end{math} is an example of this kind of similarity function. The parameter $\sigma$ of the Gaussian similarity function controls the width of the neighbourhoods. The parameter $\sigma$ acts like the parameter $\epsilon$ in the construction of the $\epsilon$-neighbourhood graph \citep{Luxburg07}.

EMG datasets provide examples of datasets where neither an $\epsilon$-neighbourhood graph, a $k$-nearest neighbour graph nor a fully connected graph can capture properly the similarities between the data instances especially when there are different densities within the same dataset. To address this issue, we propose two new similarity graph models aimed at providing greater robustness in handling different data densities within a dataset. They are referred to as probabilistic threshold graph and probabilistic criterion graph. One of the main advantages of the proposed probabilistic threshold and probabilistic criterion graphs is that they do not have a problem in dealing with instances in different scales. This means that unlike the $\epsilon$-neighbourhood graph which does not connect instances belonging to the same scale when a dataset is on different scales, and unlike the $k$-nearest neighbour graph which, in the same case, would connect instances on different scales, the proposed similarity graph models can connect instances within regions of constant density when data is on different scales. The mutual $k$-nearest neighbour graph can at times act on different scales but setting the parameter $k$ in this case usually is a problem because, first finding the optimal $k$ value for a certain dataset is tricky and second and more importantly, one dataset can have the optimal value of $k$ that does not mix the data scales with one another on one part of the dataset different from the optimal value of $k$ on another part of the same dataset.\\

In the proposed probabilistic threshold graph, a parameter $w$ is used as a threshold on the similarity values. Similarity values greater than or equal to $w$ are kept while similarity values smaller than $w$ are assigned using a truncated Gaussian distribution with mean = $w$ and standard deviation = $\sigma$. Another parameter $\mathrm{\epsilon}$ is used to decide the final similarity values as illustrated in the Probabilistic Thresholding Similarity Graph Model Section. As shown in equations~\ref{eq:one} $\&$~\ref{eq:two}, initial values of similarity, which are compared with $w$ are normalised based on the summation of distances from a certain instance. This leads to the fact that the thresholding applied here is relative to the data and does not depend on absolute values as is the case with $\epsilon$ in the $\epsilon$-neighbourhood graph and $k$ in the $k$-nearest neighbour graph. Nonetheless $w$ and $\mathrm{\epsilon}$ still provide a hard thresholding on the similarity values and therefore these parameters affect the similarity values a great deal. In order to alleviate this effect of hard thresholding, a similarity graph model that is based on a probabilistic acceptance criterion is proposed. As illustrated in the Probabilistic Acceptance Criterion Similarity Graph Model Section, similarity values are assigned either from $s_{ij} \sim N(w, \sigma)$ or to $0$ based on a probability value and there is no hard threshold under which similarity values are directly assigned a value of $0$. The proposed similarity graph models can be formally described as follows:
\subsubsection*{Probabilistic Thresholding Similarity Graph Model}\label{sec:methodsoneone}
For each vertex $v_i$, distances between a vertex $v_i$ and all other vertices; $d_{ij},j=1,...,n, j\neq i,$ are calculated first as Euclidean distances. Initial similarity values are subsequently calculated as a function of the distances by equation~\ref{eq:one}.

\begin{equation}\label{eq:one}
\mathrm{s}_{ij}^{init}=\frac{{d_{ij}}^{m}}{{\sum_{j=1}^nd_{ij}}^{m}},\;m<0\;\mbox{is a smoothing parameter}
\end{equation}

Similarity values greater than or equal to $w$ are kept, while the rest of similarity values are assigned using a truncated Gaussian distribution with mean = $w$ and standard deviation = $\sigma$. $m$ is a smoothing parameter that controls the normalised similarity values by tuning the ratio given to each distance. In the limit $m = 0$, distances are assigned equal weight and $w$ acts the same way as $k$ in the $k$-nearest neighbour similarity graph model. On the other hand, the larger the absolute value of $m$ the larger the weight assigned to the smallest distance and the smaller the weight assigned to the rest. $m = -1$ is used in all the experiments, because it keeps a convenient number of distances greater than or equal to $w$. Also, experiments were applied using $m = -1,-2,-3,-4\;\&-5$, and the highest classification accuracy has always been obtained with $m = -1$.

If $s_{ij}> w$, $s_{ij}$ is used to represent the respective edge weight on the similarity graph model. For the interval $s_{ij}\in (0,w)$, values of $s_{ij} \sim N(w, \sigma)$ are used to decide the final value of $s_{ij}$ as follows. If a weight value generated by $N(w, \sigma)$ is smaller than a certain small threshold value $\mathrm{\epsilon}$, then the respective similarity value is set to $0$, otherwise the similarity value is set to the generated weight. In summary, define $\mathrm{s_\epsilon}$ as $f(\mathrm{s}_\mathrm{\epsilon},w,\sigma) = \mathrm{\epsilon}$, where $f(\mathrm{s},w,\sigma)$ is defined in equation~\ref{eq:two}. Then, similarity values greater than or equal to $w$ are taken as they are, similarity values smaller than $\mathrm{s_\epsilon}$ are set to $0$, and for the interval $s_{ij}\in (\mathrm{s_\epsilon},w)$  similarity values are assigned by $s_{ij} \sim N(w, \sigma)$, as shown in equation~\ref{eq:two}. Similarity graph models constructed by probabilistic thresholding are referred to as probabilistic threshold graphs.
\begin{equation}\label{eq:two}
s_{ij} = \begin{cases}
\mathrm{s}_{ij}^{init} & \text{if $\mathrm{s}_{ij}^{init} \geq w$}\\
f(\mathrm{s}_{ij}^{init},w,\sigma)=\frac{1}{\sigma \sqrt{2\pi}}\mathrm{e}^{-\frac{{(\mathrm{s}_{ij}^{init}-w)}^2}{2{\sigma}^2}}  & \text{if $\epsilon \leq f(\mathrm{s}_{ij}^{init},w,\sigma) < w$}\\
0 & \text{if $f(\mathrm{s}_{ij}^{init},w,\sigma) < \epsilon$}.
\end{cases}
\end{equation}
Based on comprehensive cross-validation, the optimal values of $w$ and $\sigma$ are obtained. The parameter $\sigma$ controls the width of the neighbourhoods for instances farther than $w$. In the limit $\sigma = 0$, such neighbourhoods are assigned a weight value of $0$. The smaller the value of $\sigma$, the more sparse the similarity graph model.
%with the right w, there are not huge differences in accuracy duo to changes in \sigma
\subsubsection*{Probabilistic Acceptance Criterion Similarity Graph Model}\label{sec:methodsonetwo}
Distances and corresponding initial similarity values are calculated the same way as in thresholding. Similarity values that are greater than or equal to $w$ are again kept as they are while a truncated Gaussian distribution with mean = $w$ and standard deviation = $\sigma$ is utilised as follows in order to calculate similarity values smaller than $w$. The weight values resulting from $N(w, \sigma)$ are accepted as they are into the neighbourhood with a probability based on the generated weight and therefore a stochastic acceptance criterion, that does not require a threshold, is provided. To sum it up, similarity values greater than or equal to $w$ are taken as they are, while for interval $s_{ij}\in (0,w)$ similarity values are obtained either by $s_{ij} \sim N(w, \sigma)$, with a probability based on the weight generated from $N(w, \sigma)$, or set to 0 otherwise, as displayed in equation~\ref{eq:three}. Similarity graph models constructed by probabilistic acceptance criterion are referred to as probabilistic criterion graphs.
\begin{equation}\label{eq:three}
s_{ij} = \begin{cases}
\mathrm{s}_{ij}^{init} & \text{if $\mathrm{s}_{ij}^{init} \geq w$}\\
f(\mathrm{s}_{ij}^{init},w,\sigma) = \frac{1}{\sigma \sqrt{2\pi}}\mathrm{e}^{-\frac{{(\mathrm{s}_{ij}^{init}-w)}^2}{2{\sigma}^2}}  &
          \text{ with prob. $\propto f(\mathrm{s}_{ij}^{init},w,\sigma)$}\\
	0 & \text{with prob. $\propto 1-f(\mathrm{s}_{ij}^{init},w,\sigma)$}.
\end{cases}
\end{equation}
Both neighbourhood relationships of the proposed similarity graph models are turned into symmetric neighbourhoods in a fashion similar to the $k$-nearest neighbour graph; either by assigning the maximum value out of $similarity(v_i, v_j)\;\&$ $similarity(v_j, v_i)$ to both of them or by taking the minimum value out of these two values to be their updated symmetric similarity value.\\

\begin{figure}[!]
 \centerline{\includegraphics{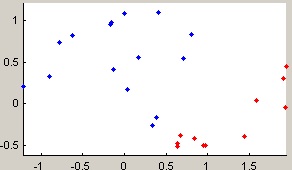}}
  \caption[DatasetA]{$DatasetA$: A 26-instance 2-label dataset is used to test different similarity graph models. Instances are coloured according to their labels, i.e.\ blue instances are labelled blue whereas red instances are labelled red.\label{fig:two}}
%\vspace{2em}
%\squeezeup
\end{figure}

Examples where the advantages of the proposed probabilistic threshold and probabilistic criterion graphs are clear, usually relate to groups that have irregular shapes. For example, Figure~\ref{fig:two} shows a toy dataset representing a pattern that takes place quite often in the EMG datasets as well as other datasets where there are two or more (two in the case of Figure~\ref{fig:two}) irregular groups in the data. A Matlab GUI, which was developed by \citet{Hein2007}, is tailored in order to show the figures used throughout this illustrative example.

\begin{figure}[!]
\centerline{\includegraphics{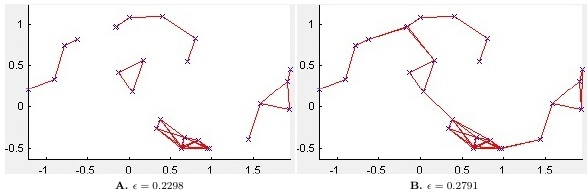}}
%\centering
%\mbox{\subfigure[$\epsilon=0.2298$]{\includegraphics[width=3in, height=1.75in]{./epsilon229804.jpg}}
%\subfigure[$\epsilon=0.2791$]{\includegraphics[width=3in, height=1.75in]{./epsilon2791.jpg} }}
%\newline
\caption{Similarity graph model of $DatasetA$ as a result of applying an $\epsilon$-neighbourhood graph. This technique fails to identify the 2 groups of $DatasetA$. \textbf{A.} $\epsilon=0.2298$ leads to $5$ groups as it loosely or never connects instances belonging to the same correct group. \textbf{B.} $\epsilon=0.2791$ leads to $1$ group as it overconnects instances belonging to the two different correct groups.\label{fig:three}}
\end{figure}

Figure~\ref{fig:three} shows how the instances are connected when an $\epsilon$-neighbourhood graph is used with values of $\epsilon$ equal to $0.2298$ and $0.2791$. When $\epsilon$ is less than the former, number of groups or connected components is $\ge$ 5 whereas number of groups is always 1 for values of $\epsilon$ greater than the latter. In Figure~\ref{fig:three}(A), $\epsilon = 0.2298$ is the value of $\epsilon$ that resulted from leave-one-out cross-validation on this small dataset and it leads to $5$ groups as it loosely or never connects instances belonging to the same correct group (a correct group refers to a group in Figure~\ref{fig:two}). Bigger values of $\epsilon$, like $\epsilon = 0.2791$ in Figure~\ref{fig:three}(B), overconnects instances belonging to the two different correct groups.

\begin{figure}[!]
\centerline{\includegraphics{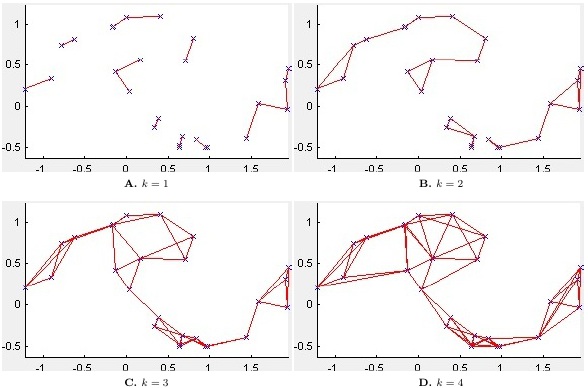}}
%\centering
%\mbox{\subfigure[$k=1$]{\includegraphics[width=3in, height=1.75in]{./symmetricKNN1.jpg}}
%\subfigure[$k=2$]{\includegraphics[width=3in, height=1.75in]{./symmetricKNN2.jpg} }}
%\newline
%\mbox{\subfigure[$k=3$]{\includegraphics[width=3in, height=1.75in]{./symmetricKNN3.jpg}}
%\subfigure[$k=4$]{\includegraphics[width=3in, height=1.75in]{./symmetricKNN4.jpg} }}
%\newline
\caption{Similarity graph model of $DatasetA$ as a result of applying a symmetric $k$-nearest neighbour graph. This technique fails to identify the 2 groups of $DatasetA$. \textbf{A.} $k=1$ leads to $9$ groups as it hardly connects more than few instances in each correct group. \textbf{B.} $k=2$ leads to $3$ groups one of them (the middle group) containing instances belonging to both correct groups. \textbf{C.} $k=3$ leads to $1$ group as it overconnects instances belonging to the two different correct groups. \textbf{D.} $k=4$ leads to $1$ group as it overconnects instances belonging to the two different correct groups.\label{fig:four}}
\end{figure}

Figure~\ref{fig:four} shows the similarity graph models when a symmetric $k$-nearest neighbour graph is used with values of $k$ equal to $1$, $2$, $3$ and $4$. The number of groups is always $1$ for values of $k$ greater than 4. No value of $k$ made a symmetric $k$-nearest neighbour graph get the correct groups. The value of $k$ resulting from leave-one-out cross-validation is $k=3$ and it connects instances belonging to the two different correct groups.

\begin{figure}[!]
\centerline{\includegraphics{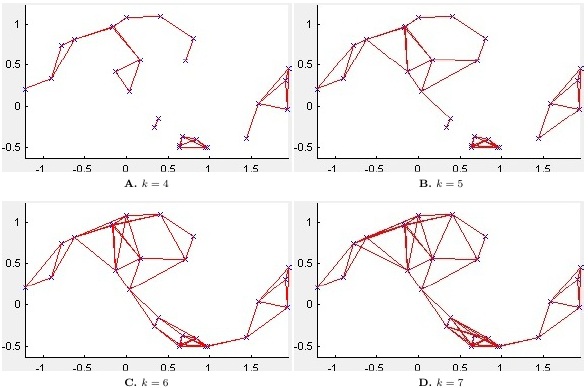}}
%\centering
%\mbox{\subfigure[$k=4$]{\includegraphics[width=3in, height=1.75in]{./mutualKnn4.jpg}}
%\subfigure[$k=5$]{\includegraphics[width=3in, height=1.75in]{./mutualKnn5.jpg} }}
%\newline
%\mbox{\subfigure[$k=6$]{\includegraphics[width=3in, height=1.75in]{./mutualKnn6.jpg}}
%\subfigure[$k=7$]{\includegraphics[width=3in, height=1.75in]{./mutualKnn7.jpg} }}
%\newline
\caption{Similarity graph model of $DatasetA$ as a result of applying a mutual $k$-nearest neighbour graph. This technique fails to identify the 2 groups of $DatasetA$. \textbf{A.} $k=4$ leads to $4$ groups as it does not connect all instances of each correct group.  \textbf{B.} $k=5$ leads to $3$ groups as it does not connect all instances of the right side correct group.  \textbf{C.} $k=6$ leads to $1$ group as it overconnects instances belonging to the two different correct groups. \textbf{D.} $k=7$ leads to $1$ group as it overconnects instances belonging to the two different correct groups.\label{fig:five}}
\end{figure}

Figure~\ref{fig:five} shows the similarity graph models when a mutual $k$-nearest neighbour graph is used with values of $k$ equal to $4$, $5$, $6$ and $7$. The number of groups is always $1$ for values of $k$ greater than 7. No value of $k$ made a mutual $k$-nearest neighbour graph get the correct groups. The value of $k$ resulting from leave-one-out cross-validation is $k=4$ and even if it is a better fit than both the $\epsilon$-neighbourhood graph and the symmetric $k$-nearest neighbour graph, the two disconnected components on the right side of Figure~\ref{fig:five}(A) should have been connected as one group because, as per Figure~\ref{fig:two}, they belong to the same correct group. The same goes for the group on the right side along with the one in the middle of Figure~\ref{fig:five}(A).

\begin{figure}[!]
 \centerline{\includegraphics{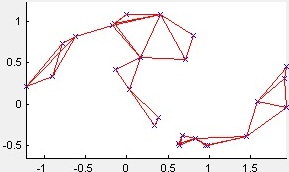}}
  \caption[labelled]{Similarity graph model of $DatasetA$ as a result of applying a probabilistic threshold graph with $w=0.073$. This similarity graph model manages to correctly identify the 2 groups of $DatasetA$ as it can connect instances belonging to different densities. It bases its decision whether or not to connect instances based on a weight value. The values that each weight value is compared to are normalised values representing the distance between a certain instance and another divided by summation of distances between the former and all instances of the dataset. This normalization leads to a similarity graph model (both the probabilistic threshold and probabilistic criterion graphs) that not only depends on absolute values of parameters but is also heavily impacted by relative weight values where a certain distance value from a certain instance to another is taken into consideration only with relative to another distance from the former instance to a third instance.\label{fig:six}}
\end{figure}

Figure~\ref{fig:six} shows the similarity graph model resulting when a probabilistic threshold graph is used with a value of weight threshold $w$ equal to $0.073$ which is the value resulting from applying leave-one-out cross-validation on this illustrative dataset. The probabilistic threshold graph is the only similarity graph model that leads to the correct groups because the values that $w$ are compared to are normalised values representing the distance between a certain instance and another divided by summation of distances between the former to all instances of the dataset. This normalization leads to a similarity graph model that not only depends on absolute values of parameters but is also heavily impacted by relative weights where a distance from a certain instance to another is taken into consideration relative to other distances from the former instance to all others.
% \citet{SabatoST10}
%and activity recognition \citep{StikicS09}.

\subsection*{Spectral Graph-Theoretic Grouping}
Formal notations of a general form of grouping and graph grouping are presented, followed by three equations (equations~\ref{eq:four},~\ref{eq:five} $\&$~\ref{eq:six}) presenting the main graph Laplacian matrices in the literature.

\paragraph{Grouping Input} The learner receives a set $X$ of $n$ i.i.d.\ instances where each instance has $p$ features. Even if it is not always the case, but let's assume another number $k$ is given, representing the number of groups. This is in line with this study.
\paragraph{Grouping Output} The learner is required to return a partition of the $n$ instances into $k$ disjoint subsets $C_1, C_2,...,C_k$, where $\bigcup_{i=1}^k C_i = X$ \citep{BD06}. A good partitioning should minimise pairwise distances among instances of the same subset and maximise pairwise distances among instances of different subsets, so that subsets are homogeneous and well separated, respectively.
\paragraph{Graph-Theoretic Grouping} The learner is required to return a partition of a graph into disjoint subsets, or groups of vertices, where edges between vertices of different groups have weights that are as low as possible (well separated groups) and edges between vertices within the same group have weights that are as high as possible (homogeneous groups).

Let $D$ be the degree matrix and $W$ be the edge weight matrix of the similarity graph.  The unnormalised graph Laplacian is equal to the following:
\begin{equation}\label{eq:four}
L = D - W
\end{equation}
There are two ways by which a normalised graph Laplacian can be calculated \citep{Chung97}, which are as follows:
\begin{equation}\label{eq:five}
L_{nor1} = D^{-\frac{1}{2}} L D^{-\frac{1}{2}}
\end{equation}
or
\begin{equation}\label{eq:six}
L_{nor2} = D^{-1} L
\end{equation}

A normalised graph Laplacian is calculated in all the spectral graph-Theoretic grouping algorithms performed in this study via equation~\ref{eq:six}. The following is a description of the spectral graph-Theoretic grouping algorithm used in the experiments:
%\paragraph{Unnormalized Spectral Clustering}
\paragraph{Normalised Spectral Graph-Theoretic Grouping according to Shi and Malik \citep{Shi2000}}
\begin{itemize}
\item Construct a similarity graph model $S \in R^{n \times n}$ as one of the models described in the Similarity Graph Models Section or one of the two introduced similarity graph models.
\item Calculate the normalised Laplacian $L_{nor2} = D^{-1} L$.
\item Calculate the first $k$ eigenvectors ${u}_1,{u}_2,...,{u}_k$ of $L_{nor2}$. $k$ is the number of groups. In our study, $k = 2$.
\item Let $U \in R^{n \times k}$ be a matrix whose columns are ${u}_1,{u}_2,...,{u}_k$
\item Let $y_i \in R^k$ be the $i^{th}$ row of $U$.
\item Group $y_i$, $i=1,2,...,n$ into subsets $C_1,C_2,...,C_k$ using k-means.
\end{itemize}
%\newpage

%%%%%%%%%%%%%%%%%%%%%%%%%%%%%%%%%%%%%%%%%%%%%%%%%%%%%%%%%%%%%%%%%%%%%%%%%%%%%%%%%%%%%%%%%%%%%%%%%%%%%%%%%%%%%%%%%%%%%%%%%%
%%%%%%%%%%%%%%%%%%%%%%%%%%%%%%%%%%%%%%%%%%%%%%%%%%%%%%%%%%%%%%%%%%%%%%%%%%%%%%%%%%%%%%%%%%%%%%%%%%%%%%%%%%%%%%%%%%%%%%%%%%
\section{Results}\label{s:results}
%Latex has no command for the conditional independence symbol
\newcommand{\Perp}{\perp \! \! \! \perp}
%F-measure		%in terms of reliability of the corresponding weak annotations
The proposed weakly supervised learning approach mainly consists of a spectral graph-theoretic grouping strategy, which in turn is based on similarity graph models, and a subsequent weak classifier. Therefore, investigating the performance of similarity graph models is crucial to the performance evaluation of the weakly supervised learning approach. Datasets used in the evaluation of the proposed weakly supervised learning approach as a whole are real-world EMG datasets, whilst datasets used in a separate evaluation of the introduced similarity graph models are three benchmark datasets, Abalone, Swiss Banknotes and Segmentation.
%Experimental results on the real-world EMG muscle datasets demonstrate that the proposed weakly supervised learning approach can provide significant improvements in classification performance. Also, experimental results on the benchmark datasets demonstrate that the proposed similarity graph models provide accurate grouping results, which should lead to reliable weak annotations.

There is no ground truth labelling available for the unlabelled instances of the EMG datasets. The main purpose of weakly annotating unlabelled instances is to improve the performance of the subsequent weak classifier. This means that accuracy of the weak classifier is the main metric for measuring the quality of the weak annotation. Still, we present the results of an internal evaluation metric of grouping in order to demonstrate the quality of the grouping strategy in a generic sense. Davies-Bouldin index is used as an internal evaluation measure for the EMG datasets.

For a $2$-group problem like the EMG grouping (see the Methods Section), Davies-Bouldin can be calculated as follows:
\begin{equation}\label{eq:seven}
%\mbox{Davies-Bouldin} = \frac{1}{2}\sum_{i=1}^2 {max}_{i \neq j}(\frac{\sigma_i + \sigma_j}{d(c_i,c_j)})
\mbox{Davies-Bouldin} = \frac{\sigma_1 + \sigma_2}{d(c_1,c_2)}
\end{equation}
where $c_i, i = 1\;or\;2,$ is the centroid of group $i$, and $\sigma_i, i = 1\;or\;2,$ is the average distance of all instances of group $i$ to centroid $c_i$, and $d(c_1,c_2)$ is the distance between the two centroids \citep{Davies79}. As the numerator expresses the compactness of the groups of a grouping result (intra-group distance) and the denominator expresses the separation among the groups (inter-group distance), the smaller the value of Davies-Bouldin index, the better the corresponding grouping. Values of Davies-Bouldin index do not depend on number of groupings nor the grouping algorithm in use \citep{Davies79}. Another advantage of Davies-Bouldin index is that it has a better time complexity than most other internal evaluation measures of grouping\citep{Petrovic06}.

Benchmark grouping datasets used in the experiments of this study have their ground truth labels available. Ground truth labels were never used in the learning process by any means. F1 score is used as an external evaluation measure for these datasets where ground truth labels are available. F1 score is a grouping external evaluation measure that weights the recall by a parameter $\beta$ \citep{Rijsbergen79}. Precision is $P = \frac{\mbox{true positives}}{\mbox{true positives}\;+\;\mbox{false positives}}$ and recall is $R = \frac{\mbox{true positives}}{\mbox{true positives}\;+\;\mbox{false negatives}}$. F1 score is calculated by:
\begin{equation}\label{eq:eight}
\mbox{F1 score} = \frac{(\beta^2+1)\;P\;R}{\beta^2 P + R}
\end{equation}
Here we use $\beta = 1$. Therefore, F1 score used here is the harmonic mean of precision and recall:
\begin{equation}\label{eq:nine}
\mbox{F1 score} = \frac{2\;P\;R}{P + R}
\end{equation}
Best value of F1 score is $1$ or $100\%$ and worst value is $0$. In this range, the larger the value of F1 score, the better the corresponding grouping result.

Results are divided into spectral graph-theoretic results, which are based on similarity graph models, and weakly supervised classification results. The former represent the content of the Analysis of Similarity Graph Models Section which provides an analysis of similarity graph models, two of which are proposed in this study. The second part of the results compares between weakly supervised classifiers and the corresponding fully supervised classifier (the Weakly Supervised Classifier vs.\ Fully Supervised Classifier Section).

\subsection*{Analysis of Similarity Graph Models}\label{sec:resultsone}
Regarding the EMG datasets, the parts used of every EMG dataset in the similarity graph models represent the unlabelled parts. The outcome of each spectral graph-theoretic grouping consists of two groups. The labelled part of every EMG dataset is used to annotate the groups because the group with a smaller number of elements is assigned the same label (normal) as the labelled instances while the group with more elements is assigned the disordered label. This assumption is based on the structure of a disordered muscle, which typically contains more disordered MUs than normal MUs. The elements which get annotated by the spectral grouping represent weakly labelled data that can be later used as further annotated training data for classification.

For the other three datasets, it is an ordinary spectral grouping problem. Grouping is applied on all the data as there is no labelled part of the data. Spectral grouping is evaluated on its own in this section and then its impact on classification performance is evaluated in the Weakly Supervised Classifier vs.\ Fully Supervised Classifier Section. As a part of the spectral graph-theoretic grouping evaluation, the proposed probabilistic threshold and probabilistic criterion similarity graph models (see the Methods Section) are evaluated based on grouping evaluation measures by comparing them to other similarity graph models existent in the literature.

In Table~\ref{tab:one}, values of the evaluation measures are shown for the following datasets:
\begin{itemize}
	\item Abalone \citep{Blake98} dataset: 4177 instances, 9 features in 10 groups.
	\item Swiss Banknotes \citep{Flury83} dataset: 200 instances, 6 features in 2 groups.
	%\item Chain Link \citep{Ultsch05} dataset: 1000 instances, 3 features in 2 groups.
	\item Segmentation \citep{Likas03} dataset: 2310 instances, 19 features in 7 groups.
	\item EMG Myopathic Upper Leg dataset (Myo Upper Leg): 557 instances, 8 features in 2 groups.
	\item EMG Neurogenic Upper Leg dataset (Neuro Upper Leg): 356 instances, 8 features in 2 groups.
	\item EMG Myopathic Lower Leg dataset  (Myo Lower Leg): 583 instances, 8 features in 2 groups.
	\item EMG Neurogenic Lower Leg dataset  (Neuro Lower Leg): 444 instances, 8 features in 2 groups.
\end{itemize}

The similarity graph models experimented are:
\begin{itemize}
	\item Probabilistic threshold: optimal values of $w$ and $\sigma$ are obtained by cross-validation (see the Methods Section for a detailed illustration of the probabilistic threshold graph). There are two versions of the probabilistic threshold graph as per how to transform the similarity matrix into a symmetric matrix:
		\begin{itemize}
			\item Prob.\ threshold Min.: Assign the minimum value out of $similarity(v_i, v_j)\;\&\;similarity(v_j, v_i)$ to both of them
			\item Prob.\ threshold Max: Assign the maximum value out of $similarity(v_i, v_j)\;\&\;similarity(v_j, v_i)$ to both of them
		\end{itemize}
	\item Probabilistic criterion: optimal values of $w$ and $\sigma$ are obtained by cross-validation (see the Methods Section for a detailed illustration of the probabilistic criterion graph). There are two versions of the probabilistic criterion graph as per how to transform the similarity matrix into a symmetric matrix:
		\begin{itemize}
			\item Prob.\ acceptance Min.: Assign the minimum value out of $similarity(v_i, v_j)\;\&\;similarity(v_j, v_i)$ to both of them
			\item Prob.\ acceptance Max: Assign the maximum value out of $similarity(v_i, v_j)\;\&\;similarity(v_j, v_i)$ to both of them
		\end{itemize}
	\item $\epsilon$-neighbourhood: optimal value of $\epsilon$ is obtained by cross-validation.
	\item $k$-nearest neighbour: optimal value of $k$ is obtained by cross-validation.
	\item Mutual $k$-nearest neighbour: optimal value of $k$ is obtained by cross-validation.
	\item Fully connected graph: optimal value of $\sigma$ is obtained by cross-validation.
\end{itemize}

The first 3 datasets are publicly available datasets that have been used in grouping before. The latter 4 datasets represent real EMG datasets. These datasets were acquired from upper leg and lower leg recordings and each of them contains two types of instances; normal as well as disordered (myopathic or neurogenic). In fact there are 2 rather than 4 EMG datasets as the myopathic and neurogenic Upper Leg datasets represent bags of the same dataset (the same goes for the Lower Leg dataset) but they are shown as two different datasets here because they are treated separately as far as spectral graph-theoretic grouping and its evaluation are concerned. In the Weakly Supervised Classifier vs.\ Fully Supervised Classifier Section, where weakly supervised classification is applied, weakly annotated data of both Upper Leg datasets are being processed together along with the strongly labelled instances of the Upper Leg (resp.\ Lower Leg) dataset. All EMG data were collected under IRB approval and were de-identified.

\begin{figure}[!]
 \centerline{\includegraphics[width=100mm,height=75mm]{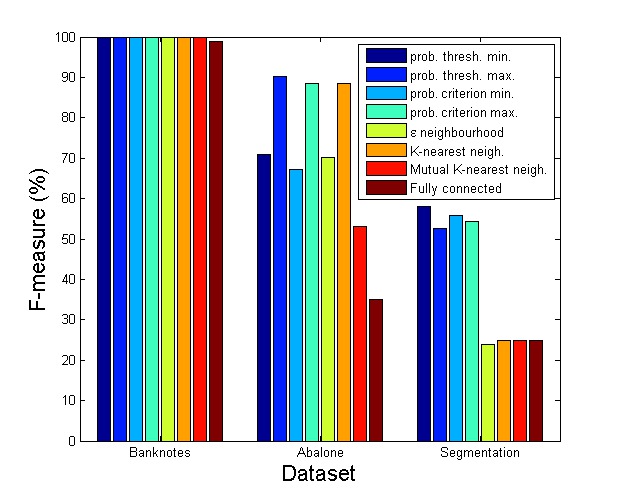}}
  \caption[F1 score values for datasets with ground truth labels]{F1 score values for datasets with ground truth labels. The greater the F1 score value the better. Grouping results are either better (Abalone and Segmentation datasets) or on a par with the best results (Banknotes dataset), with the proposed probabilistic threshold and probabilistic criterion graphs.\label{fig:seven}}
\end{figure}

\begin{figure}[!]
 \centerline{\includegraphics[width=110mm,height=85mm]{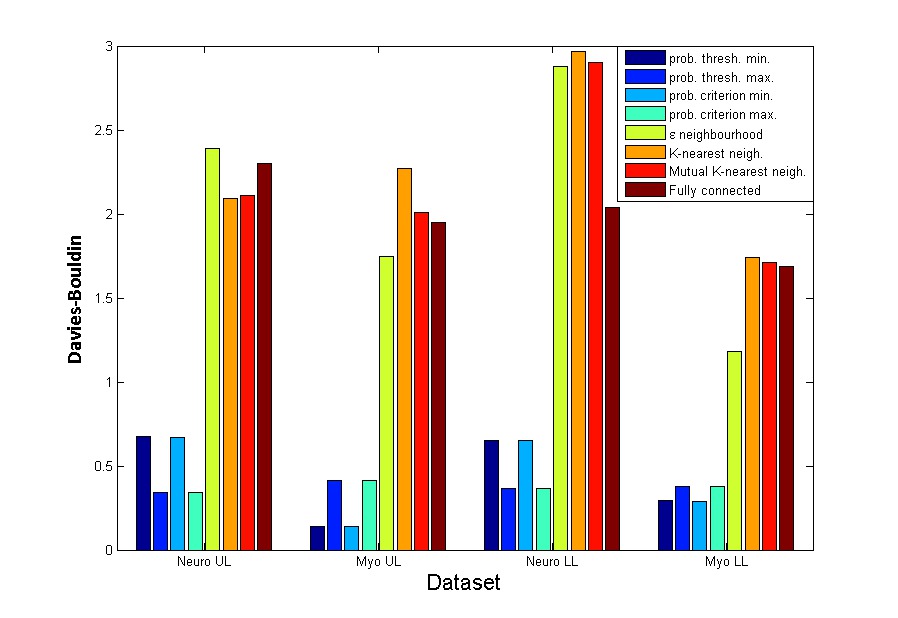}}
  \caption[Davies-Bouldin index values for datasets without ground truth labels]{Davies-Bouldin index values for datasets without ground truth labels. The smaller the index value the better. Grouping results are by far better on all EMG datasets with the proposed probabilistic threshold and probabilistic criterion graphs.\label{fig:eight}}
\end{figure}

As mentioned earlier, for the F1 score, the greater the value the more accurate the similarity graph model, while for Davies-Bouldin index, the smaller the value the better (higher similarity within a group and lower similarity between groups) the similarity graph model. As can be seen in Table~\ref{tab:one} and in Figure~\ref{fig:seven} and Figure~\ref{fig:eight}, with the proposed probabilistic similarity graph models, grouping results are better, or at least as good as, the other similarity graph models. Figure~\ref{fig:eight} shows the improvement achieved by using any of the four proposed similarity graph models (minimum and maximum graphs of probabilistic threshold, and minimum and maximum graphs of probabilistic criterion) over the other similarity graph models in comparison as the Davies-Bouldin index values are clearly better with the former. The same conclusion is shown in case of the Segmentation dataset (by far the largest out of the 3 datasets) in Figure~\ref{fig:seven}. For the Abalone dataset in Figure~\ref{fig:seven}, the probabilistic threshold maximum graph leads to the best result as its F1 score value is slightly better than the one achieved by constructing the probabilistic criterion maximum graph as well as the K-nearest neighbour graph. All graphs are nearly equally good for the Banknotes dataset displayed in Figure~\ref{fig:seven}. One other advantage of the proposed similarity graph models lies in the fact they do not depend on distance among the instances, location of the instances nor on a number of neighbours specified a priori that can perform well at some part of the dataset but not on another part of the same dataset due to, for example, having a dataset containing different densities within it.

Results show that both the probabilistic threshold and the probabilistic criterion graphs lead to very similar grouping results among themselves as shown by the values of the validity indices. The former leads to better results in case of the Abalone dataset and the minimum graph of the Segmentation dataset, while the latter leads to slightly better results in case of the maximum graph of the Segmentation dataset. For the rest of the datasets, results are quite similar among the two proposed similarity graph models.

On the other hand, for some datasets minimum similarity graph models lead to better results than maximum similarity graph models and vice versa on the other datasets. This suggests that the choice between the minimum and maximum similarity graph models for the same model (probabilistic threshold or probabilistic criterion) should depend on the value of the group validity index resulting from cross-validation.

\subsection*{Weakly Supervised Classifier vs.\ Fully Supervised Classifier}\label{sec:resultstwo}
Now that the annotated training data is larger due to the addition of weakly labelled instances that got annotated by spectral grouping, we want to evaluate the significance of the weak labelling procedure with respect to classification performance on the EMG datasets. Results of the classification are evaluated before and after weak labelling, i.e.\ with a fully supervised classifier and with a weak classifier that uses weakly labelled data resulting from spectral grouping along with strongly labelled data.

Results are presented using three different classifiers. The classifications algorithms used are logistic regression, $k$-nearest neighbour and Quadratic Discriminant Analysis (QDA).

\paragraph{Logistic Regression}
Assuming $F$ refers to a $p$-dimensional feature vector of an instance $I$, the utilised logistic regression classifier assumes the following:
\begin{equation}\label{eq:ten}
P(I=i|F) \propto \exp^{\beta_{i[0]} + \beta_{i[1:p]} f}, i < 3.
\end{equation}

\paragraph{$k$-Nearest Neighbour}
$k$-nearest neighbour classifiers usually perform well in cases where the decision boundary is complex and there is enough data to train. Here, the value of $K$ is obtained by cross-validation.

\paragraph{Quadratic Discriminant Analysis (QDA)}
QDA assumes a Gaussian distribution of each label and assigns to an instance the label with a greater posterior probability. Parameters of the Gaussian distribution of each class are estimated from the training data by a maximum likelihood estimate (MLE). A posterior probability refers to $p(I|F)$, and the assigned label is obtained by $i = argmax_i\;p(F|I)(p(I)$.\\

In the fully supervised classifier, unlabelled instances are dealt with as follows. An instance belonging to a myopathic or a neurogenic bag is assumed to be myopathic or neurogenic respectively. This assumption is not accurate because myopathic and neurogenic bags contain normal instances. Therefore, it is easy to see that performance of the fully supervised classifier would severely suffer due to this assumption, which is demonstrated by the results. On the other hand, the weak classifier exploits weak labels assigned to each of the previously unlabelled instances by the spectral grouping procedure. The Upper Leg dataset has $650$ labelled instances and $913$ unlabelled instances. The fully supervised classifier assigns the respective bag label to each unlabelled instance. On the other hand, the weakly supervised classifier processes $913$ weakly labelled instances as well as $650$ strongly labelled instances. The Lower Leg dataset has $672$ labelled instances and $1027$ unlabelled instances. Therefore, the weakly supervised classifier processes $1027$ weakly labelled instances as well as $672$ strongly labelled instances.

Leave-one-out cross-validation is implemented by setting instances belonging to a single muscle as test data while training on instances of the rest of the muscles, then repeating this process for every muscle. It can be more precisely referred to as leave-one-muscle-out cross-validation as far as this study is concerned. Overall muscle classification accuracy is the main metric used to evaluate the classification performance.

Table~\ref{tab:two} shows the results of the fully supervised and weakly supervised classifiers. Every weak classifier is named after the spectral graph-theoretic grouping strategy pursued but three different classification algorithms are utilised with each. Results show that, using a weak classifier, muscle classification accuracy significantly improves compared to the fully supervised classifier. Logistic regression performs slightly better compared to both k-nearest neighbours and QDA, but the difference is not huge. This shows that if the training instance annotation process is properly performed, the classification results are stable and not algorithm-dependent.

%%%%%%%%%%%%%%%%%%%%%%%%%%%%%%%%%%%%%%%%%%%%%%%%%%%%%%%%%%%%%%%%%%%%%%%%%%%%%%%%%%%%%%%%%%%%%%%%%%%%%%%%%%%%%%%%%%%%%%%%%%
\section{Discussion}\label{s:discussion}

A weakly supervised learning paradigm is introduced. The goal is to improve classification performance by first weakly annotating unlabelled samples of a training dataset using a spectral graph-theoretic grouping strategy, then using the weakly annotated data along with the strongly labelled data to construct a larger annotated training set to be used in classification. Spectral graph-theoretic grouping exploits similarity among data instances as well as the relationship between unlabelled and strongly labelled data instances, by constructing similarity graph models to weakly annotate unlabelled data instances. Two new similarity graph models, which provide greater robustness in handling different data densities within a dataset, are introduced. Afterwards, a classifier learns from the weakly as well as strongly labelled data. Results show that performance of the resulting weakly supervised classifier as a whole is better than its counterpart fully supervised classifier on the EMG datasets. Also, results of experiments performed on benchmark and EMG datasets show that the spectral graph-theoretic grouping strategy based on the introduced similarity graph models leads to grouping results better than or on a par with similarity graph models in the literature.  The proposed spectral graph-theoretic grouping strategy for weakly supervised learning provided improved performance compared to its fully supervised learning counterpart primarily due to the fact that such an approach can obtain a reliable set of weakly labelled data that, when augmented with strongly labeled data to form a larger annotated training dataset, can be used to train a classifier with stronger classification performance than can be achieved using only strongly labelled data, or strongly labelled data and unlabelled data.  Furthermore, the proposed similarity graph models led to improved results due primarily to the flexibility of the models that can adapt to the underlying data when compared to existing graph models which require more strict and require more rigid parameter optimization.

%%%%%%%%%%%%%%%%%%%%%%%%%%%%%%%%%%%%%%%%%%%%%%%%%%%%%%%%%%%%%%%%%%%%%%%%%%%%%%%%%%%%%%%%%%%%%%%%%%%%%%%%%%%%%%%%%%%%%%%%%%

%[ht] COMMENTED FOR NOW

%%%%%%%%%%%%%%%%%%%%%%%%%%%%%%%%%%%%%%%%%%%%%%%%%%%%%%%%%%%%%%%%%%%%%%%%%%%%%%%%%%%%%%%%%%%%%%%%%%%%%%%%%%%%%%%%%%%%%%%%%%

%\newpage

% \subsubsection*{Acknowledgements}
%
% Use unnumbered third level headings for the acknowledgements title.
% All acknowledgements go at the end of the paper.

\renewcommand{\refname}{}
\subsubsection*{References}
\vspace{-1.5em}
\small

%\bibliography{spectral_bib}

\renewcommand{\bibnumfmt}[1]{#1.}

\section*{Acknowledgements}
This research was undertaken, in part, thanks to funding from the Canada Research Chairs program. The study was also funded by the Natural Sciences and Engineering Research Council (NSERC) of Canada and the Ontario Ministry of Economic Development and Innovation.

\section*{Author Contributions}
TA and AW were involved in designing the study and performing performance analysis. TA and AW were involved in the writing and editing. DS was involved in the editing. All authors reviewed the manuscript.

\section*{Additional Information}
Competing financial interest: All authors in this study have no competing financial interests.
\bibliographystyle{/home/folkert/scripts/vancouver}

\renewcommand{\thesubfigure}{\textbf{\Alph{subfigure}.}}	%\renewcommand\thesubfigure{\roman{subfigure}}


\begin{table*}[!]
\caption{Grouping indices values based on different similarity graph models.\label{tab:one}}
%\vspace{-5em}
\begin{center}
\small
\begin{tabular}{|c|c|c|c|}
\hline
%Upper Leg & \multicolumn{5}{c||}{:} & \multicolumn{4}{c||}{: } & Non-MIL\\
\textbf{Dataset} & \textbf{Similarity Graph Model} & \textbf{Eval. Index} & \textbf{Value}\\
\hline
\multirow{8}{*}{\textbf{Abalone}} 	& Prob.\ threshold Min & \multirow{8}{*}{F1 score} & 71$\%$\\
						& Prob.\ threshold Max & & 90.3$\%$\\
						& Prob.\ criterion Min & & 67.2$\%$\\
						& Prob.\ criterion Max & & 88.37$\%$\\
						& $\epsilon$-neighbourhood & & 70.1$\%$\\
						& $k$-nearest neighbour & & 88.6$\%$\\
						& Mutual $k$-nearest neighbour & & 53.2$\%$\\
						& Fully connected graph & & 35.1$\%$\\
\hline
\multirow{8}{*}{\textbf{Banknotes}} 	& Prob.\ threshold Min & \multirow{8}{*}{F1 score} & 100$\%$\\
						& Prob.\ threshold Max & & 100$\%$\\
						& Prob.\ criterion Min & & 100$\%$\\
						& Prob.\ criterion Max & & 100$\%$\\
						& $\epsilon$-neighbourhood & & 100$\%$\\
						& $k$-nearest neighbour & & 100$\%$\\
						& Mutual $k$-nearest neighbour & & 100$\%$\\
						& Fully connected graph & & 99$\%$\\

\hline
%\multirow{8}{*}{\textbf{Chain Link}} & Prob.\ threshold Min & \multirow{8}{*}{F1 score} & 100$\%$\\
%						& Prob.\ threshold Max & & 100$\%$\\
%						& Prob.\ criterion Min & & 90.3$\%$\\
%						& Prob.\ criterion Max & & 90.3$\%$\\
%						& $\epsilon$-neighbourhood & & 100$\%$\\
%						& $k$-nearest neighbour & & 100$\%$\\
%						& Mutual $k$-nearest neighbour & & 100$\%$\\
%						& Fully connected graph & & 99$\%$\\
%\hline
\multirow{8}{*}{\textbf{Segmentation}} & Prob.\ threshold Min & \multirow{8}{*}{F1 score} & 58.1$\%$\\
						  & Prob.\ threshold Max & & 52.5$\%$\\
%the next has been changed 49.83 (how many special people change)
						& Prob.\ criterion Min & & 55.83$\%$\\
%the next has been changed 58.6 (how many special people change)
						& Prob.\ criterion Max & & 54.4$\%$\\
						& $\epsilon$-neighbourhood & & 24$\%$\\
						& $k$-nearest neighbour & & 24.97$\%$\\
						& Mutual $k$-nearest neighbour & & 24.95$\%$\\
						& Fully connected graph & & 24.95$\%$\\
\hline
\multirow{8}{*}{\textbf{Myo Upper Leg}} 	& Prob.\ threshold Min & \multirow{8}{*}{Davies-Bouldin index} & 0.1402\\
						& Prob.\ threshold Max & & 0.4137\\
						& Prob.\ criterion Min & & 0.1400\\
						& Prob.\ criterion Max & & 0.4132\\
						& $\epsilon$-neighbourhood & & 1.75\\
						& $k$-nearest neighbour & & 2.27\\
						& Mutual $k$-nearest neighbour & & 2.01\\
						& Fully connected graph & & 1.95\\
\hline
\multirow{8}{*}{\textbf{Neuro Upper Leg}} 	& Prob.\ threshold Min & \multirow{8}{*}{Davies-Bouldin index} & 0.6727\\
						& Prob.\ threshold Max & & 0.3413\\
						& Prob.\ criterion Min & & 0.6725\\
						& Prob.\ criterion Max & & 0.3411\\
						& $\epsilon$-neighbourhood & & 2.39\\
						& $k$-nearest neighbour & & 2.09\\
						& Mutual $k$-nearest neighbour & & 2.11\\
						& Fully connected graph & & 2.3\\
\hline
\multirow{8}{*}{\textbf{Myo Lower Leg}} 	& Prob.\ threshold Min & \multirow{8}{*}{Davies-Bouldin index} & 0.2918\\
						& Prob.\ threshold Max & & 0.3807\\
						& Prob.\ criterion Min & & 0.2911\\
						& Prob.\ criterion Max & & 0.3806\\
						& $\epsilon$-neighbourhood & & 1.18\\
						& $k$-nearest neighbour & & 1.74\\
						& Mutual $k$-nearest neighbour & & 1.71\\
						& Fully connected graph & & 1.69\\
\hline
\multirow{8}{*}{\textbf{Neuro Lower Leg}} 	& Prob.\ threshold Min & \multirow{8}{*}{Davies-Bouldin index} & 0.6521\\
						& Prob.\ threshold Max & & 0.3642\\
						& Prob.\ criterion Min & & 0.6520\\
						& Prob.\ criterion Max & & 0.3641\\
						& $\epsilon$-neighbourhood & & 2.88\\
						& $k$-nearest neighbour & & 2.97\\
						& Mutual $k$-nearest neighbour & & 2.9\\
						& Fully connected graph & & 2.04\\
\hline
\end{tabular}
\end{center}
\end{table*}
\normalsize

\makeatletter
\setlength{\@fptop}{0pt}
\makeatother
\begin{table*}[ht!]
\caption{Muscle classification accuracy based on the proposed weakly supervised classifiers vs. a fully supervised classifier.\label{tab:two}}
\vspace{0.25em}
\begin{center}
\small
\begin{tabular}{|c|c|c|c|c|}
\hline
\multirow{2}{*}{\textbf{Dataset}} & \multirow{2}{*}{\textbf{Similarity Graph Model}} &  \multicolumn{3}{c|}{\textbf{Classification Accuracy}}\\ \cline{3-5}
& & \textbf{Logistic Regression} & \textbf{K-Nearest Neighbours} & \textbf{QDA}\\
\hline
\multirow{5}{*}{\textbf{Upper Leg}} 	& Prob.\ threshold Min Weak Class.& $95\%$ & $94.1\%$ & $94\%$\\
						& Prob.\ threshold Max Weak Class.& $92.4\%$ & $92.2\%$ & $91\%$\\
						& Prob.\ acceptance Min Weak Class.& $95\%$ & $94.4\%$ & $95\%$\\
						& Prob.\ acceptance Max Weak Class.& $92.7\%$ & $92.8\%$ & $90.5\%$\\
						& Fully Supervised Class.& $81\%$ & $78.7\%$ & $79.9\%$\\
\hline
%\multirow{5}{*}{\textbf{Neuro UL}} 	& Prob.\ threshold Min Weak Class.& $93.7\%$\\
%						& Prob.\ threshold Max Weak Class.& $95.5\%$\\
%						& Prob.\ acceptance Min Weak Class.& $94\%$\\
%						& Prob.\ acceptance Max Weak Class.& $94.9\%$\\
%						& Fully Supervised Class.& $79\%$\\
%\hline
\multirow{5}{*}{\textbf{ Lower Leg}} 	& Prob.\ threshold Min Weak Class.& $96.1\%$ & $95.3\%$ & $94.9\%$\\
						& Prob.\ threshold Max Weak Class.& $93.4\%$ & $93.2\%$ & $90.7\%$\\
						& Prob.\ acceptance Min Weak Class.& $95.6\%$ & $95\%$ & $92.8\%$\\
						& Prob.\ acceptance Max Weak Class.& $93.4\%$ & $93.5\%$ & $91\%$\\
						& Fully Supervised Class.& $82.3\%$ & $77.1\%$ & $78\%$\\
\hline
%\multirow{5}{*}{\textbf{LL}} 	& Prob.\ threshold Min Weak Class.&$94.2\%$ \\
%						& Prob.\ threshold Max Weak Class.& $95.3\%$\\
%						& Prob.\ acceptance Min Weak Class.& $95\%$\\
%						& Prob.\ acceptance Max Weak Class.& $95.6\%$\\
%						& Fully Supervised Class.& $83.1\%$\\
%\hline
\end{tabular}
\end{center}
\end{table*}
\normalsize

\end{document}